\documentclass[11pt]{article}

\usepackage{acl}
\usepackage{times}
\usepackage{latexsym}
\usepackage{amssymb}

\usepackage{CJKutf8}

\usepackage{booktabs}
\usepackage{multirow}
\usepackage{graphicx}
\usepackage{xcolor}
\usepackage{arydshln}

\usepackage{algorithm}
\usepackage{algpseudocode}

\usepackage{amsmath}
\usepackage[T1]{fontenc}

\usepackage[utf8]{inputenc}

\usepackage{microtype}

\usepackage{inconsolata}

\usepackage{graphicx}

\title{Align, Integrate, and Fire: Efficient Token-Level Alignment for Zero-Shot SpeechLLMs}

\author{Abderrahmane Issam \qquad {\bf Yusuf Can Semerci} \qquad {\bf Jan Scholtes} \qquad {\bf Gerasimos Spanakis} \\
        Department of Advanced Computing Sciences \\ 
        Maastricht University \\ 
        \small{\texttt{\{abderrahmane.issam, y.semerci, j.scholtes, jerry.spanakis\}@maastrichtuniversity.nl}}}

\begin{document}
\maketitle
\begin{abstract}
While Large Language Models excel in natural language processing, efficiently extending their capabilities to spoken input remains a significant challenge. Existing methods for building SpeechLLMs often rely on computationally expensive full-model fine-tuning, or employ parameter-efficient projectors that suffer from inefficient token sequence lengths and costly full-model supervision. In this paper, we introduce Aligned Continuous Integrate-and-Fire, a highly efficient framework for zero-shot speech processing. Our method dynamically compresses continuous acoustic frames into the exact discrete token length of the target text utilizing explicit Dynamic Time Warping alignments. This allows our initial training stage to establish a robust acoustic-to-semantic bridge using lightweight distance metrics, entirely bypassing the computationally expensive LLM forward pass. For subsequent fine-tuning, we propose a memory-efficient knowledge distillation objective that targets a single LLM layer, performing competitively with full-model cross-entropy training at a fraction of the computational cost. Through extensive evaluations on Automatic Speech Recognition and Speech Translation, we demonstrate that our method achieves superior performance compared to prior parameter-efficient baselines.\footnote{Our code: \url{https://github.com/issam9/ACIF}}
\end{abstract}

\section{Introduction}

Large Language Models (LLMs) have achieved unprecedented success in natural language processing \cite{openai2024gpt4ocard, comanici2025gemini25pushingfrontier, grattafiori2024llama3herdmodels, qwen2025qwen25technicalreport, Guo_2025}, yet a significant portion of high-performing open-weights LLMs remain text-only. While cascaded systems (speech recognition model followed by an LLM) offer a straightforward way to process spoken input, they suffer from error propagation and prevent end-to-end downstream fine-tuning. To natively equip LLMs with auditory capabilities, recent works connect a speech encoder to the LLM and fine-tune the system on speech-to-text data \cite{wu2023, gong2024listen, tang2024salmonn, fathullah2024, hu-etal-2024-wavllm, das2025speechverselargescalegeneralizableaudio}. Although effective, this paradigm is computationally expensive and prone to task-specific overfitting \cite{tang2024salmonn}. 

To improve efficiency, recent approaches freeze the LLM and train only a speech projector to map acoustic features to text embeddings \cite{fathullah-etal-2024-audiochatllama, deng-etal-2025-wav2prompt, tan-etal-2025-ssr, Mohapatra2026}. However, these methods still face significant limitations. Wav2Prompt \cite{deng-etal-2025-wav2prompt} dynamically matches token lengths using a Continuous Integrate-and-Fire (CIF) module \cite{dong2020} but relies on expensive Cross Entropy (CE) supervision from the full LLM. SSR \citet{tan-etal-2025-ssr} requires external alignment tools and extensive distillation training followed by full LLM fine-tuning. SpeechMapper \cite{Mohapatra2026} avoids LLM forward passes but pads text embeddings to match the longer length of speech sequences, wasting computational resources on uninformative tokens and hindering both training and inference efficiency.

To overcome these limitations, we introduce \textbf{ACIF} (Aligned Continuous Integrate-and-Fire), a highly efficient framework for zero-shot speech processing that is trained exclusively on Automatic Speech Recognition (ASR) data. ACIF dynamically compresses continuous acoustic frames into the exact discrete token length of the target text. During our initial training phase (Stage 1), we supervise the CIF module using explicit token-level alignments generated via Dynamic Time Warping (DTW) \cite{sakoe1978-dtw} directly from the speech and text embeddings. This allows the projector to learn a robust acoustic-to-semantic mapping through lightweight distance metrics (Mean Squared Error and cosine distance), bypassing the computationally expensive LLM forward pass.

For Stage 2 fine-tuning, standard full-model CE training on ASR data risks overfitting and incurs heavy computational costs. To address this, we propose a memory-efficient Knowledge Distillation (KD) objective that distills hidden states from a single LLM layer, performing competitively with full-model CE at a fraction of the cost. Through an analysis of KD layer depth, we reveal that ASR relies heavily on shallow, localized representations optimally captured at the first layer, whereas Speech Translation (ST) benefits from the abstract semantic representations developed in intermediate layers.

We evaluate ACIF on Automatic Speech Recognition (LibriSpeech, VoxPopuli, FLEURS) and zero-shot Speech Translation (Europarl-ST, CoVoST-2) using Llama 3.1 (8B) and Qwen 2.5 (7B) backbones. Our main contributions are:
\begin{itemize}
    \item We propose ACIF, combining a CIF module with DTW-guided supervision to project continuous speech frames into exact LLM token representations.
    \item We demonstrate that our Stage 1 training establishes a high-quality semantic bridge with vastly greater efficiency than existing baselines, requiring no LLM forward passes.
    \item We introduce a highly memory-efficient, single-layer KD objective for Stage 2 fine-tuning that competes robustly with full-model CE training.
    \item We achieve superior zero-shot performance compared to prior methods, offering a substantially more lightweight and computationally efficient alternative for developing SpeechLLMs.
\end{itemize}

\begin{figure*}[t]
    \centering
    \includegraphics[width=\linewidth]{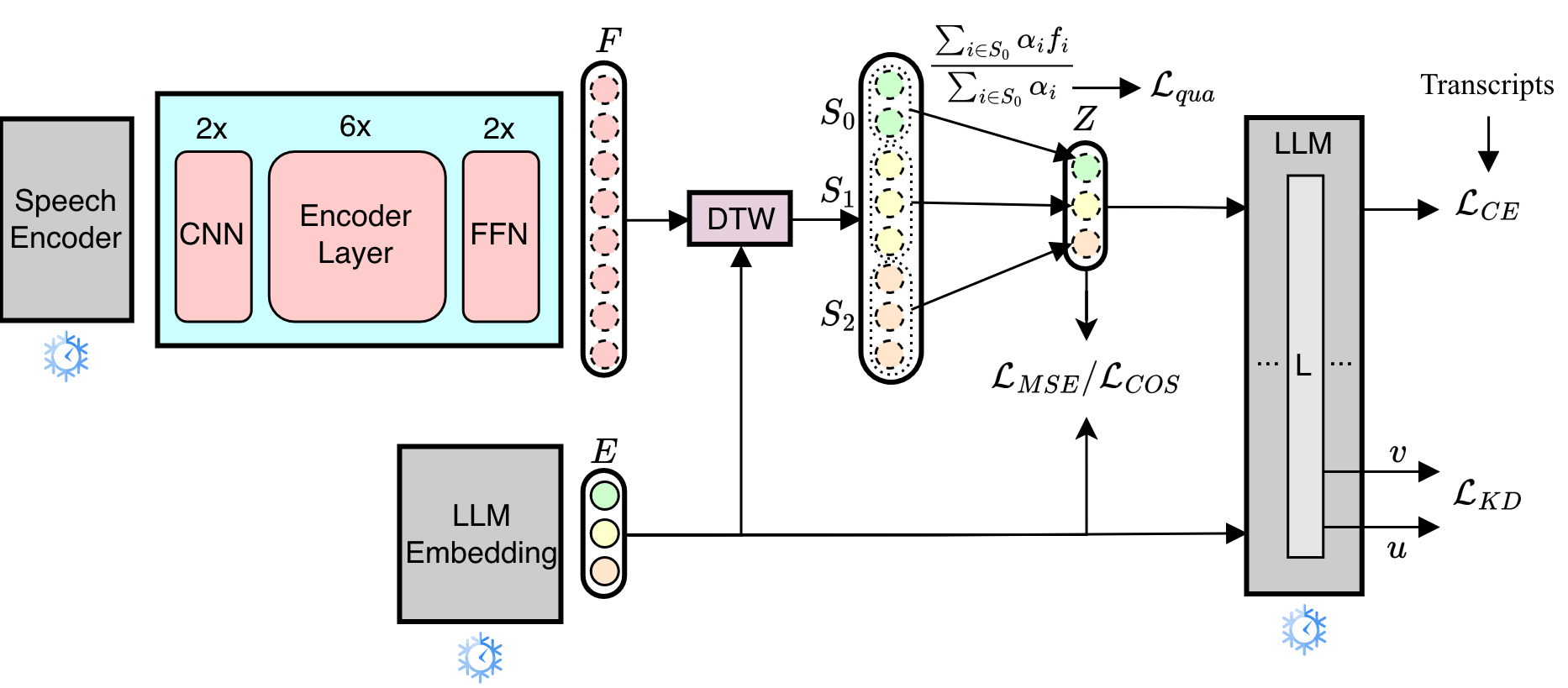}
    \caption{\textbf{Overview of the ACIF architecture.} The proposed projector consists of two convolutional layers, a 6-layer Transformer encoder, and a two-layer feed-forward network. The projected speech frames $F$ are aligned with the target text embeddings $E$ using DTW. An additional layer predicts continuous weights $\alpha_i$ to integrate each frame $f_i$ into its corresponding aligned token (Eq.~\ref{eq:alpha}). The resulting token-level speech embeddings $Z$ are optimized using normalized MSE ($\mathcal{L}_{\text{MSE}}$, Eq.~\ref{eq:mse}) and mean-centered cosine distance ($\mathcal{L}_{\text{cos}}$, Eq.~\ref{eq:cos}). Simultaneously, the $\alpha_i$ values are supervised via a quantity loss ($\mathcal{L}_{\text{qua}}$, Eq.~\ref{eq:qua}) to ensure they sum to 1 over each text token's segment. During the fine-tuning stage, we incorporate either a standard cross-entropy loss ($\mathcal{L}_{\text{CE}}$) or a knowledge distillation loss ($\mathcal{L}_{\text{KD}}$), the latter being computed between the intermediate hidden states of the text and speech embeddings extracted at layer $L$ of the truncated LLM.}
    \label{fig:method}
\end{figure*}

\section{Related Works}

\subsection{Speech Large Language Models}
While recent LLMs natively support multi-modal inputs \cite{openai2024gpt4ocard, comanici2025gemini25pushingfrontier}, a large portion of modern LLMs remain text-only, making the development of efficient audio adapters crucial. Previous works have equipped LLMs with auditory capabilities by connecting a speech encoder and fine-tuning the full model or utilizing LoRA adapters on task-specific datasets \cite{wu2023, gong2024listen, tang2024salmonn, fathullah2024, hu-etal-2024-wavllm, das2025speechverselargescalegeneralizableaudio}. Recent parameter-efficient approaches have moved beyond this by training only a lightweight projector and keeping the LLM entirely frozen \cite{fathullah-etal-2024-audiochatllama, deng-etal-2025-wav2prompt, tan-etal-2025-ssr, Mohapatra2026}. Closely related to our approach, \citet{deng-etal-2025-wav2prompt} introduced Wav2Prompt, which converts speech frames to text tokens using a CIF mechanism \cite{dong2020}. Although this successfully avoids updating the LLM parameters, it still relies on CE supervision from the LLM during training, introducing significant computational overhead. Alternatively, \citet{Mohapatra2026} proposed SpeechMapper, which makes the LLM forward pass optional. However, SpeechMapper resolves the length mismatch between modalities by introducing padding tokens to the text embeddings, which renders training and inference inefficient. In contrast, our work trains a CIF module to project speech frames to the exact text token length. Furthermore, unlike Wav2Prompt which requires full LLM propagation for training, we supervise this projection using explicit token-level alignments generated via DTW.

\subsection{Bridging the Modality Gap between Speech and Text}
The goal of aligning acoustic and semantic representations is widely studied under the framework of bridging the modality gap. In speech-to-text tasks, representation misalignment has been shown to severely bottleneck performance \cite{liu2020bridgingmodalitygapspeechtotext, wang2020-bridging}. Various alignment strategies have proven effective to alleviate this issue \cite{inaguma-etal-2021-source, ye-etal-2022-cross, fang-etal-2022-stemm, ouyang-etal-2023-waco, zhang-etal-2025-representation}, with techniques relying on explicit word- or token-level boundaries yielding particularly strong results \cite{fang-etal-2022-stemm, ouyang-etal-2023-waco, issam-etal-2025-dtw}. Moving beyond static, pre-computed alignments, \citet{issam-etal-2025-dtw} utilized DTW \cite{sakoe1978-dtw} to dynamically generate alignments during the training of end-to-end speech translation models. We similarly rely on DTW to establish correspondence between acoustic and text representations. However, rather than training a task-specific encoder-decoder architecture with simple mean pooling, we project speech into the embedding space of a frozen LLM to specifically target zero-shot generalization. Additionally, we utilize trainable CIF weights to dynamically pool and integrate the continuous speech frames into exact, discrete token representations during training and inference.

\section{Methodology}

In this section, we present our methodology for efficiently bridging a pretrained speech encoder with a frozen LLM, as illustrated in Figure~\ref{fig:method}. We first introduce the ACIF projector and its training objectives (Section~\ref{sec:acif}). To further optimize the projector, Section~\ref{sec:finetuning} details a memory-efficient fine-tuning strategy. Finally, Section~\ref{sec:inference} outlines how the trained projector dynamically integrates speech embeddings during inference.

\subsection{ACIF: Aligned Continuous Integrate and Fire}
\label{sec:acif}

Connecting a pretrained speech encoder to a frozen LLM requires a bridging projector that resolves the cross-modal gap. To enable the direct injection of speech representations into the LLM context, the projector must map acoustic features to the LLM embedding dimension, compress dense speech frame sequences to match the exact length of textual token sequences, and align the projected frames directly into the semantic embedding space of the LLM. To achieve this multi-step mapping efficiently, we design a lightweight projector architecture coupled with a dynamic alignment strategy.

Our projector consists of two 1D Convolutional Neural Network layers that perform initial temporal downsampling, and a 6-layer encoder that learns semantic alignment from speech to text. This is followed by two linear layers that map the downsampled features into the LLM hidden dimension, producing intermediate speech frame representations $F = [f_1, f_2, \dots, f_{N'}] \in \mathbb{R}^{N' \times d}$, where $N'$ is the reduced sequence length. Concurrently, a single linear layer operates on these frames to predict a scalar weight $\alpha_i \in [0, 1]$ for each frame, representing its relative contribution toward forming a discrete token embedding. 

During training, we compute an optimal alignment path between the intermediate frames $F$ and the target text token embeddings $E = [e_1, e_2, \dots, e_M] \in \mathbb{R}^{M \times d}$ using DTW-Align \cite{issam-etal-2025-dtw} based on cosine distance. DTW partitions the $N'$ frames into $M$ contiguous segments, assigning a set of frames $S_j$ to each target token $j$. Instead of uniformly averaging the segment embeddings as in standard DTW-Align, we calculate a weighted average using the predicted alpha scores. For each target token $j$, the pooled representation $z_j$ is computed by multiplying the aligned frame embeddings by their corresponding scalar weights and dividing the accumulated result by the sum of those weights:
\begin{equation}
z_j = \frac{\sum_{i \in S_j} \alpha_i f_i}{\sum_{i \in S_j} \alpha_i}
\label{eq:alpha}
\end{equation}
This token-level integration receives explicit positional signals from the alignment bounds itself, entirely eliminating the requirement for LLM forward passes to regularize the token compression.

The projector is optimized end-to-end using a joint objective comprising normalized MSE, mean-centered cosine distance, and a quantity loss. To ensure the MSE remains stable across different LLM backbones with varying embedding magnitudes, we divide the raw error by the mean squared magnitude of the target LLM embeddings. Given the pooled speech embeddings $Z = [z_1, \dots, z_M]$ and target text embeddings $E = [e_1, \dots, e_M]$, the normalized MSE is computed as:
\begin{equation}
\mathcal{L}_{\text{MSE}} = \frac{\sum_{j=1}^M \|z_j - e_j\|_2^2}{\sum_{j=1}^M \|e_j\|_2^2}
\label{eq:mse}
\end{equation}
Furthermore, raw cosine similarity between embeddings can be inflated by a shared mean vector \cite{mu2018allbutthetop}, which might dominate the pairwise similarity and obscure the semantic signal we aim to optimize for. We mitigate this by mean-centering the embeddings prior to computing cosine similarity, subtracting the mean of the target embeddings, $\mu_E = \frac{1}{M}\sum_{j=1}^M e_j$, from both the predicted and target vectors. This removes the dominant shared direction and yields a cosine loss that is more sensitive to true semantic discrepancies:
\begin{equation}
\mathcal{L}_{\text{cos}} = \frac{1}{M} \sum_{j=1}^M \Big( 1 - \cos(z_j - \mu_E, e_j - \mu_E) \Big)
\label{eq:cos}
\end{equation}
where $\cos(\cdot, \cdot)$ denotes the cosine similarity operator.

Finally, the quantity loss applies an $L_1$ penalty between the accumulated scalar weights for each token and a target value of 1, enforcing that the predicted weights correctly match the exact target sequence length:
\begin{equation}
\mathcal{L}_{\text{qua}} = \frac{1}{M} \sum_{j=1}^M \left| \sum_{i \in S_j} \alpha_i - 1 \right|
\label{eq:qua}
\end{equation}
The overall objective is the weighted sum of these three components, $\mathcal{L}_{\text{total}} = \lambda_{\text{mse}} \mathcal{L}_{\text{MSE}} + \lambda_{\text{cos}} \mathcal{L}_{\text{cos}} + \lambda_{\text{qua}} \mathcal{L}_{\text{qua}}$.

\subsection{Efficient Fine-tuning}
\label{sec:finetuning}
Our projector can be efficiently fine-tuned using feedback from the LLM to further improve cross-modal embedding alignment. While the LLM itself could be jointly fine-tuned for downstream speech-to-text tasks, we freeze the LLM to preserve its general capabilities and prioritize computational efficiency. To fine-tune the projector, a standard approach is to incorporate a Cross Entropy loss alongside the alignment objectives \cite{deng-etal-2025-wav2prompt, Mohapatra2026}:
\begin{equation}
\mathcal{L}_{\text{total}} = \lambda_{\text{mse}} \mathcal{L}_{\text{MSE}} + \lambda_{\text{cos}} \mathcal{L}_{\text{cos}} + \lambda_{\text{qua}} \mathcal{L}_{\text{qua}} + \lambda_{\text{ce}} \mathcal{L}_{\text{CE}}
\end{equation}
where $\lambda_{\text{ce}}$ governs the weight of the autoregressive language modeling loss.

However, while optimizing only the projector minimizes the number of trainable parameters, computing $\mathcal{L}_{\text{CE}}$ still requires a full forward pass through the multibillion-parameter LLM. This imposes a severe GPU memory bottleneck during training. We hypothesize that for zero-shot performance, the primary requirement is successfully mapping speech into the LLM's initial semantic space; thus, full-depth LLM propagation is unnecessary, and shallow feedback is sufficient to guide the representations. 

Guided by this assumption, we propose an efficient alternative to standard CE loss by utilizing early-layer LLM representations (e.g., $1$ layer) for fine-tuning. To enforce deeper semantic alignment without the full memory overhead, we introduce a KD objective utilizing the logit lens technique \cite{nostalgebraist2020logitlens}. 

Specifically, the pretrained LLM is structurally pruned during training to retain only the first $L$ transformer layers. Both the target text embeddings $E$ and the pooled speech embeddings $Z$ are passed through this truncated network to extract their respective intermediate hidden states. These states are then mapped directly into the vocabulary space by applying the LLM's unembedding layer, yielding teacher text logits $\mathbf{u}$ and student speech logits $\mathbf{v}$. 

The distillation loss $\mathcal{L}_{\text{KD}}$ is computed as the average Kullback-Leibler (KL) divergence between the teacher and student probability distributions over the vocabulary:
\begin{equation}
\mathcal{L}_{\text{KD}} = \frac{1}{M} \sum_{j=1}^M D_{\text{KL}}\Big(\text{softmax}(\mathbf{u}_j) \parallel \text{softmax}(\mathbf{v}_j)\Big)
\label{l{KD}}
\end{equation}
where $M$ is the number of valid target tokens. In configurations where memory efficiency is paramount, $\mathcal{L}_{\text{KD}}$ safely replaces $\mathcal{L}_{\text{CE}}$, regularizing the projector by forcing the intermediate speech representations to mirror the predictive trajectory of the text embeddings at a fraction of the computational cost.

\subsection{Inference Stage}
\label{sec:inference}
At inference time, since target transcriptions are unavailable and DTW alignment cannot be applied, the CIF module dynamically collapses the variable-length sequence of speech frame embeddings into discrete token representations. It sequentially aggregates frame embeddings scaled by their predicted scalar weights, progressively accumulating acoustic evidence until the predefined firing threshold of 1 is reached. This threshold signifies the completion of a semantic token. To prevent information loss at token boundaries, the specific frame that breaches the threshold is fractionally partitioned: a remainder portion completes the current token embedding, while the surplus is carried over to initialize the subsequent token. Once all frames are processed, the module resolves any residual acoustic information left in the tail buffer. If the accumulated tail weight exceeds a half-token threshold ($w \ge 0.5$), it is deemed semantically meaningful and emitted as a valid final token. Crucially, this trailing embedding is divided by its accumulated weight to yield a weighted mean, normalizing its magnitude to match that of fully integrated tokens. 

\section{Experiments}
\subsection{Datasets}
For model training, we exclusively utilize the LibriSpeech dataset \cite{Panayotov2015}, leveraging its paired audio and text transcriptions. We subsequently evaluate ASR performance on the standard LibriSpeech, Voxpopuli \cite{wang-etal-2021-voxpopuli} and FLEURS \cite{fleurs2022arxiv} test splits. To assess zero-shot ST capabilities, we report results on select test sets from the Europarl-ST \cite{iranzo2020} and CoVoST-2 \cite{wang21s_interspeech} benchmarks. Specifically, we evaluate on the English-to-Spanish, German, Italian, and French directions from Europarl-ST, alongside the English-to-German and Chinese directions from CoVoST-2. Although our proposed alignment method is language-agnostic and applicable to any source language with sufficient ASR data, we restrict our evaluation to English-centric pairs to ensure direct comparison with prior work. 

\subsection{Backbones}
For our acoustic representations, we use SeamlessM4T-v2-Large \cite{communication2023seamlessmultilingualexpressivestreaming} as a speech encoder. Following SpeechMapper \cite{Mohapatra2026}, we specifically extract features from the 24th layer of the SeamlessM4T encoder. For the core language model, we experiment with Llama-3.1-8B-Instruct \cite{grattafiori2024llama3herdmodels} and Qwen2.5-7B-Instruct \cite{qwen2025qwen25technicalreport}. To preserve their pretrained capabilities and ensure computational efficiency, the weights of both the speech encoders and the LLMs remain strictly frozen during training.

\subsection{Training Details}

To optimize GPU memory and accelerate the training process, we pre-compute the acoustic features from the frozen speech encoder. Consequently, only the trainable projector and the frozen LLM embedding layer need to be loaded into memory. Target text transcriptions are normalized using the MMS normalization \cite{Pratap2024}. Specifically, we remove HTML tags, apply NFKC normalization, lowercase, and remove punctuation. Architecturally, the projector consists of two 1D convolutional layers (kernel size 5, stride 2), which reduce the temporal sequence length by a factor of 4. These are followed by 6 standard Transformer encoder layers with a hidden dimension of 1024. A final two-layer feed-forward network projects the representations through an intermediate dimension of 2048 into the specific LLM hidden dimension (4096 for Llama-3.1-8B and 3584 for Qwen2.5-7B). 

We optimize the model using AdamW with a peak learning rate of $1 \times 10^{-4}$ and a cosine learning rate scheduler, incorporating a 10\% linear warmup period. Training utilizes dynamic batching with a maximum batch size of 8192 speech frames and runs for 250k steps. This effectively matches the total volume of training data seen in SpeechMapper \cite{Mohapatra2026}, adjusting for their comparatively smaller batch size. During the subsequent fine-tuning stage, we maintain identical hyperparameters but reduce the learning rate to $1 \times 10^{-5}$ and the batch size to 1024 tokens and train for 20k steps. We strictly evaluate the final saved checkpoint in both stages. 

During the initial training phase, the weights for the loss components are set to $\lambda_{\text{mse}} = 0.01$, $\lambda_{\text{cos}} = 10.0$, and $\lambda_{\text{qua}} = 1.0$.\footnote{The disparity between $\lambda_{\text{cos}}$ and $\lambda_{\text{mse}}$ balances their initial gradient magnitudes: while the cosine loss is naturally bounded in $[0, 2]$, MSE between the randomly initialized projector output and the pretrained LLM embedding space is initially much larger and would otherwise dominate early optimization.} For the subsequent fine-tuning stage, we maintain these base hyperparameters while introducing the additional semantic objective: either standard cross-entropy ($\mathcal{L}_{\text{CE}}$) with $\lambda_{\text{ce}} = 1.0$, or knowledge distillation ($\mathcal{L}_{\text{KD}}$) with $\lambda_{\text{kd}} = 1.0$. When utilizing $\mathcal{L}_{\text{KD}}$, the pretrained LLM is pruned to retain only its first transformer layer.

Our models are implemented using PyTorch Lightning\footnote{\url{https://github.com/Lightning-AI/pytorch-lightning}}. Due to the lightweight nature of the projector—comprising 96.6M trainable parameters—and our memory-efficient design, the entire training pipeline fits easily on a single 40GB A100 GPU. This low memory footprint even permits the concurrent training of multiple models on the same hardware. Training completes in approximately 6h30min on a single A100 GPU (or 10h30min on an NVIDIA RTX A5000), representing a drastic reduction in computational overhead compared to prior methods like SpeechMapper, which requires 4 days across 4 V100 GPUs.

\subsection{Evaluation}
For ASR evaluation, we report the Word Error Rate (WER) and Character Error Rate (CER) following the application of MMS text normalization \cite{Pratap2024}. For Speech Translation, we utilize the COMET\footnote{Unbabel/wmt22-comet-da} metric \cite{rei-etal-2022-comet}, which evaluates the generated hypothesis against both the source transcription and the reference translation. Additionally, we report case-sensitive, detokenized BLEU \cite{papineni2002} computed via SacreBLEU \cite{post-2018-call} in our code repository. During the evaluation of the CoVoST-2 dataset, we filter out audio samples containing fewer than 1,000 or more than 480,000 acoustic frames. To ensure a fair comparison, we adopt closely identical prompt templates to SpeechMapper \cite{Mohapatra2026} (see Appendix \ref{sec:prompt}). All text generation is performed using greedy decoding with a maximum generation length of 150 tokens, implemented via the HuggingFace Transformers library \cite{wolf-etal-2020-transformers}.
 
\subsection{Baselines}
Our primary baseline is SpeechMapper \cite{Mohapatra2026}, which similarly employs a two-stage training pipeline to achieve zero-shot speech capabilities. Their speech projector is architecturally heavier, consisting of two sequential blocks, each containing a single CNN layer and a 6-layer Transformer encoder. Within this structure, a linear layer in the first block projects the acoustic features to a hidden dimension of 2048, and a subsequent linear layer in the second block projects them to 4096, before a final linear layer maps the representations into the LLM's input space. Consequently, the SpeechMapper projector contains 277M trainable parameters, making it nearly three times larger than our lightweight 96.6M-parameter module. For a direct and fair comparison, we benchmark exclusively against their zero-shot experimental setup. This pipeline consists of Stage 1, which trains the projector independently without any forward passes through the LLM, and Stage 2, which incorporates the frozen LLM to fine-tune the projector using a cross-entropy objective.

\section{Results}
\subsection{ASR Performance}

Table \ref{tab:asr_results} presents the ASR performance evaluated via WER and CER on the LibriSpeech (test-clean, test-other), VoxPopuli (VP), and FLEURS datasets. Our proposed ACIF alignment framework demonstrates significant improvements over the SpeechMapper baseline across multiple dimensions. 

Most notably, our method exhibits vastly superior sample efficiency and alignment quality during the initial modality-bridging phase. Using the identical Llama 3.1 backbone, our Stage 1 model achieves a WER of 5.1 on the LibriSpeech clean split, nearly halving the 9.4 WER reported by SpeechMapper's Stage 1. This massive performance gap extends to the more challenging out-of-domain VP and FLEURS datasets, where our Stage 1 model improves upon SpeechMapper's Stage 1 WER by absolute margins of 8.8 and 13.0 points, respectively. This indicates that our alignment-guided CIF module establishes a highly accurate semantic mapping purely from the acoustic representations, without requiring expensive forward passes through the LLM. 

Furthermore, during the Stage 2 fine-tuning phase, our KD objective consistently outperforms standard CE. For Llama 3.1, KD training yields a WER of 4.1, surpassing SpeechMapper's fully fine-tuned Stage 2 result (4.9 WER) on LibriSpeech test-clean. The stabilizing effect of the KD loss becomes especially critical when employing the Qwen 2.5 backbone. While standard CE fine-tuning leads to a noticeable degradation in performance compared to its Stage 1 baseline (e.g., WER increasing from 3.7 to 4.9 on test-clean), the KD objective successfully prevents this. Notably, applying KD safely maintains the performance reached during Qwen 2.5's Stage 1, yielding the best overall CERs on both VP (9.1) and FLEURS (10.1). Ultimately, these findings demonstrate that employing KD with only a single LLM layer is a highly robust, memory-efficient alternative to full-model CE supervision.

Finally, our best-performing configuration surpasses the zero-shot baseline of SSR \cite{tan-etal-2025-ssr} on both LibriSpeech evaluation splits. Specifically, our ACIF (Stage 2) [KD] model with a Qwen-2.5 backbone achieves 3.8 and 7.2 WER on test-clean and test-other, respectively, outperforming SSR's 5.0 and 8.1.

\begin{table}[t]
    \scriptsize
    \centering
    \setlength{\tabcolsep}{3pt}
    \begin{tabular}{@{}ll ccccc@{}}
        \toprule
        & & \textbf{LS clean} & \textbf{LS other} & \textbf{VP} & \textbf{Fleurs}\\
        \midrule
        \multicolumn{2}{l}{Seamless ASR} & \textbf{2.7 / 0.9} & \textbf{5.1 / 2.0} & \textbf{8.9 / 6.2} & \textbf{8.1 / 4.7} \\
        \midrule
        
        \multirow{2}{*}{\rotatebox[origin=c]{90}{\scriptsize Llama 3.1}} 
        & SpeechMapper \scriptsize{Stage 1} & 9.4 / 6.5 & 12.0 / 7.9 & 25.0 / 19.7 & 30.2 / 27.6 \\
        & SpeechMapper \scriptsize{Stage 2} & 4.9 / 2.7 & 7.8 / 4.1 & \underline{14.8} / 9.2 & \underline{16.6} / 11.1  \\[4ex]
        \midrule
        
        \multirow{3}{*}{\rotatebox[origin=c]{90}{\scriptsize Llama 3.1}} 
        & \rule{0pt}{3ex}ACIF (stage 1) & 5.1 / 3.2 & 9.1 / 5.8 & 16.2 / 11.5 & 17.2 / 12.1  \\
        & ACIF (stage 2) [CE] & 4.7 / 2.8 & 7.9 / 4.4 & 16.3 / 11.4 & 18.3 / 11.7 \\
        & ACIF (stage 2) [KD] & 4.1 / 2.2 & 8.3 / 4.8 & \underline{14.8} / 9.7 & 17.4 / 11.7 \\[1ex]

        \midrule
        \multirow{3}{*}{\rotatebox[origin=c]{90}{\scriptsize Qwen 2.5}} 
        & \rule{0pt}{3ex}ACIF (stage 1) & \underline{3.7} / \underline{1.7} & 7.3 / \underline{3.7}  & 15.2 / 9.2  & 17.9 / 10.4 \\
        & ACIF (stage 2) [CE] & 4.9 / 2.5 & 9.1 / 4.8 & 18.3 / 12.0 & 21.6 / 13.5 \\
        & ACIF (stage 2) [KD] & 3.8 / 1.8 & \underline{7.2} / \underline{3.7} & 15.0 / \underline{9.1} & 17.2 / \underline{10.1} \\[1ex]
        
        \bottomrule
    \end{tabular}
    \caption{ASR performance (WER / CER) on LibriSpeech, VoxPopuli (VP), and FLEURS. The overall best results are highlighted in bold, while the highest SpeechLLM results are underlined. Baseline results for SpeechMapper are taken directly from the original paper \cite{Mohapatra2026}.}
    \label{tab:asr_results}
\end{table}

\subsection{Zero-Shot Speech Translation Performance}

Table \ref{tab:st_results} presents the zero-shot ST COMET scores on the Europarl-ST and CoVoST-2 evaluation sets. Consistent with our ASR findings, our ACIF method significantly outperforms the SpeechMapper baseline. Using the Llama 3.1 backbone, our Stage 1 model achieves substantial gains over SpeechMapper's Stage 1 across nearly all evaluation directions, with the exception of En-It. This improvement is particularly pronounced on the CoVoST-2 dataset, where our method outperforms the baseline by $+12.1$ COMET points on En-De and $+11.9$ points on En-Zh. Furthermore, while SpeechMapper suffers from performance degradation during its Stage 2 fine-tuning, our Stage 2 training consistently enhances the learned representations across both LLM backbones. 

When comparing the Stage 2 optimization strategies, the results indicate that both CE and KD provide robust semantic alignment for zero-shot ST. For the Llama 3.1 backbone, KD yields the highest zero-shot performance on En-Es, En-Fr, and En-It directions, peaking at a COMET score of 79.1 on En-Es. Conversely, standard CE fine-tuning proves highly effective on the CoVoST-2 dataset, achieving the top zero-shot scores of 81.1 on En-De and 83.0 on En-Zh. Strikingly, unlike the representation degradation observed during Qwen 2.5 ASR fine-tuning, both CE and KD successfully stabilize and even slightly improve upon the Qwen 2.5 Stage 1 translation baseline. Ultimately, our best zero-shot configurations establish a highly effective semantic bridge. They closely trail the performance of the explicitly supervised cascaded baselines and consistently outperform the in-domain Seamless ST model on the Europarl-ST directions without requiring any translated text during training.

Finally, in comparison to Wav2prompt \cite{deng-etal-2025-wav2prompt}, which only reports BLEU on En–Es and En–Fr, we re-evaluate ACIF (Stage 2) [KD] with Llama 3.1 backbone using their exact decoding configuration (beam size of 5 and repetition penalty of 1.5). Under these settings, our method achieves 26.7 BLEU on En–Es and 21.0 BLEU on En–Fr, compared to 25.1 and 21.7, respectively. Notably, our approach performs competitively without requiring full LLM forward propagation or the 10 hours of in-domain Europarl-ST supervision utilized in Wav2prompt's ASR stage.

\begin{table*}[t]
    \scriptsize
    \centering
    \newcommand{\std}[1]{{\color{gray}\scriptsize$\pm$#1}}
    \newcommand{\annot}[1]{{\color{darkgray}\scriptsize (#1)}}
    \setlength{\aboverulesep}{0.9pt}
    \setlength{\belowrulesep}{0pt}
    \begin{tabular}{@{}ll cccc | cc@{}}
        \toprule
        & &  \multicolumn{4}{c}{Europarl} & \multicolumn{2}{c}{CoVoST2} \\
        \cmidrule(lr){3-6} \cmidrule(lr){7-8}
        & & \textbf{en-es} & \textbf{en-fr} & \textbf{en-de} & \textbf{en-it} & \textbf{en-de} & \textbf{en-zh} \\
        \midrule
        \multicolumn{2}{l}{Transcripts + Qwen 2.5 7B \annot{topline}} & 83.53 & 79.87 & 80.55 & 79.2 & 83.3 & \textbf{87.3} \\
        \multicolumn{2}{l}{Transcripts + Llama 3.1 8B \annot{topline}} & \textbf{84.4} & \textbf{80.92} & \textbf{82.87} & \textbf{81.03} & \textbf{85.9} & 87.1 \\
        \midrule
        \multicolumn{2}{l}{Seamless + Qwen 2.5 7B \annot{Cascaded}} & 81.3 & 77.6 & 78.3 & 77.6 & 81.4 & 85.8 \\
        \multicolumn{2}{l}{Seamless + Llama 3.1 8B \annot{Cascaded}} & 82.2 & 78.7 & 80.9 & 79.1 & 84.1 & 85.6 \\
        \midrule
        \multicolumn{2}{l}{Seamless ST \annot{in-domain}} & 78.7 & 72.5 & 68.6 & 72.8 & 86.0 & 83.7 \\
        \midrule
        \multirow{2}{*}{\rotatebox[origin=c]{90}{\scriptsize Llama 3.1}} &
        SpeechMapper (stage 1) \annot{zero-shot} & 76.4 & 73.9 & 72.3 & 76.8 & 67.1 & 69.3 \\
         & SpeechMapper (stage 2) \annot{zero-shot} & 74.7\std{2.7} & 71.0\std{2.8} & 66.4\std{2.6} & 73.2\std{2.6} & 63.7\std{1.0} & 68.6\std{1.5} \\ [4ex]
        \midrule
        \multirow{3}{*}{\rotatebox[origin=c]{90}{\scriptsize Llama 3.1}} 
        & ACIF (stage 1) \annot{zero-shot} & 78.3 & 75.7 & 77.8 & 76.1 & 79.2 & 81.2 \\
        & ACIF (stage 2) [CE] \annot{zero-shot} & 78.9 & 76.2 & \underline{78.6} & 76.9 & \underline{81.1} & \underline{83.0} \\
        & ACIF (stage 2) [KD] \annot{zero-shot} & \underline{79.1} & \underline{76.3} & 78.4 & \underline{77.1} & 80.9 & 82.8 \\[2ex]
        \cdashline{1-8}[2pt/2pt] 
        \multirow{3}{*}{\rotatebox[origin=c]{90}{\scriptsize Qwen 2.5}} 
        & \rule{0pt}{3ex}ACIF (stage 1) \annot{zero-shot} & 76.6 & 73.3 & 75.8 & 73.9 & 78.6 & 82.4 \\
        & ACIF (stage 2) [CE] \annot{zero-shot} & 77.4 & 73.8 & 75.9 & 74.1 & 78.3 & 82.9 \\
        & ACIF (stage 2) [KD] \annot{zero-shot} & 77.0 & 73.7 & 76.2 & 74.3 & 78.6 & 82.4 \\[1ex]
        \bottomrule
    \end{tabular}
    \caption{Speech translation performance (COMET scores) on the Europarl-ST and CoVoST-2 datasets. The overall best results are highlighted in bold, while the highest-performing zero-shot configurations are underlined. Baseline results for SpeechMapper are taken directly from the original paper \cite{Mohapatra2026}.}
    \label{tab:st_results}
\end{table*}

\section{Analysis}
\subsection{Some Embeddings are Easier to Align}
\begin{figure}[t]
    \centering
    \includegraphics[width=\linewidth]{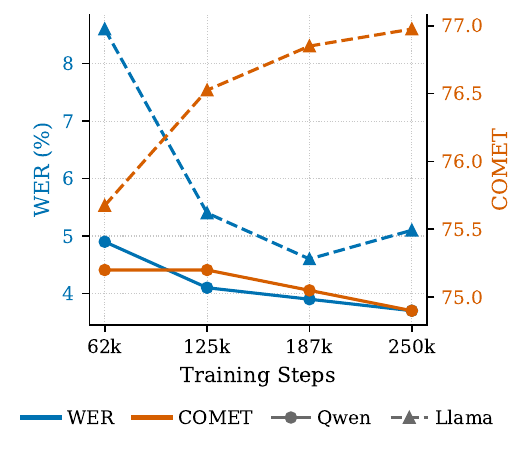}
    \caption{Impact of projector training duration on Librispeech test-clean (WER) and average Europarl-ST (COMET) performance. The Qwen 2.5 backbone achieves optimal alignment rapidly, while the performance of Llama 3.1 continues to steadily improve.}
    \label{fig:scaling}
\end{figure}

The geometry of the target embedding space may influence how easily a speech projector can learn the cross-modal mapping. Consequently, training the projector with identical hyperparameters across different LLM backbones may be suboptimal. In Figure~\ref{fig:scaling}, we evaluate the impact of training duration by varying the number of training steps for both Llama 3.1 and Qwen 2.5 backbones. We report WER on the LibriSpeech test-clean split and average COMET scores across Europarl-ST directions. The results highlight a clear difference in learning dynamics between the two models. While the projector aligned with Qwen achieves most of its performance early in training (within a quarter of the total steps), Llama steadily improves over longer training durations. This indicates that Qwen's target space is learned much faster by the projector.

Prior work has shown that pretrained language representations exhibit highly anisotropic structure and considerable redundancy, with much of the variance concentrated in a relatively small number of principal directions \cite{mu2018allbutthetop, raunak-etal-2019-effective}. Such low effective dimensionality has been argued to simplify optimization in downstream learning tasks \cite{aghajanyan-etal-2021-intrinsic}. Motivated by these observations, we examine whether differences in the geometry of the target embedding space can explain the projector learning dynamics observed above.

\begin{figure}[t]
    \centering
    \includegraphics[width=\linewidth]{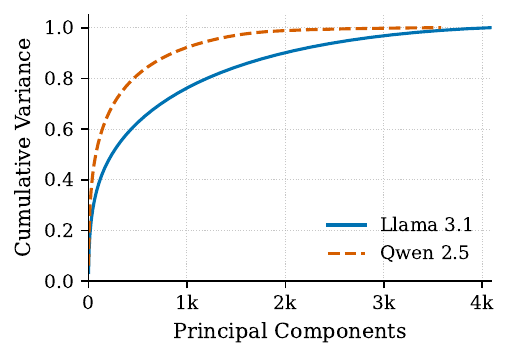}
    \caption{PCA eigenvalue decay of the target embedding spaces for Llama 3.1 and Qwen 2.5. Qwen 2.5 exhibits a substantially steeper decay, capturing 90\% of the total variance in 857 principal components compared to 1,987 for Llama 3.1.}
    \label{fig:pca_decay}
\end{figure}

Figure~\ref{fig:pca_decay} shows the cumulative variance explained by the principal components for both models, computed from over one million token embeddings extracted from the LibriSpeech training set. Qwen 2.5 exhibits a noticeably steeper eigenvalue decay than Llama 3.1. Specifically, Qwen requires only 857 principal components to explain 90\% of the variance, whereas Llama requires 1,987 (more than twice as many), indicating that the variance of Qwen embeddings is concentrated in considerably fewer directions.

These geometric differences offer a plausible explanation for the training efficiency observed in Figure~\ref{fig:scaling}. The more concentrated variance spectrum of the Qwen embedding space suggests that its target representations are easier for the speech projector to learn, which is consistent with the substantially faster convergence observed during training. These results suggest that projector optimization is influenced not only by model size or architecture, but also by the geometry of the target embedding space. Consequently, training schedules and optimization hyperparameters should be adapted to the effective complexity of the target representation rather than transferred unchanged across LLM backbones.

\subsection{Impact of Distillation Depth}
\begin{figure}[t]
    \centering
    \includegraphics[width=\linewidth]{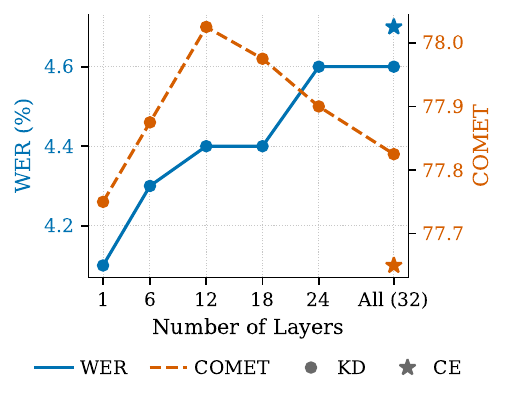}
    \caption{Impact of knowledge distillation layer depth on ASR and ST performance of Llama 3.1. WER on LibriSpeech (test-clean) steadily degrades as the number of distillation layers increases. Conversely, average COMET scores on Europarl-ST initially improve, reaching a peak at 12 layers before declining.}
    \label{fig:kd_depth}
\end{figure}

Our initial results demonstrate that KD using only the first layer of the LLM performs comparably to standard CE training with the full model. This raises an important question: how does increasing the distillation depth affect downstream performance? Figure~\ref{fig:kd_depth} plots the WER on LibriSpeech and the average COMET score on Europarl-ST as we vary the target LLM layer $L$ for the $\mathcal{L}_{\text{KD}}$ objective. The isolated star markers represent the baseline performance using standard CE supervision across the full 32-layer LLM.

The results reveal a stark contrast between the optimal representations for ASR and ST. For ASR (the solid blue line), performance is strictly optimized at the shallowest depth. Distilling only the first LLM layer yields the lowest WER, and as the distillation depth increases, acoustic alignment steadily degrades. This suggests that the ASR task relies more heavily on the localized, surface-level representations found in early layers rather than the highly abstract representations formed deeper in the network.

Conversely, ST performance (the dashed orange line) benefits from intermediate semantic abstraction. Average COMET scores initially improve as the distillation depth increases, peaking at layer 12 before steadily declining. This indicates that the ST task requires the more complex semantic signals that develop at intermediate stages of the model. Importantly, for both tasks, distilling shallow or intermediate representations yields strictly better performance than mimicking the final layers (All 32). Furthermore, our optimal KD configurations (layer 1 for ASR, layer 12 for ST) outperform the full-model CE baselines. This confirms our hypothesis that mapping speech into the LLM's early or intermediate semantic spaces is sufficient for strong zero-shot performance, eliminating the need for full-model feedback. Finally, in Appendix~\ref{sec:cosine}, we show that using cosine distance as an alternative to KL divergence is equally effective for zero-shot ASR and ST.

\subsection{Sequence Length Alignment and Efficiency}
\label{seq_length}
During inference, ACIF relies solely on the predicted accumulation weights $\alpha$ to integrate acoustic frames into discrete token representations. This raises the question of how closely the length of these integrated embeddings matches that of the target reference text. As reported in Table~\ref{tab:length_matching_results}, our ACIF mechanism dynamically aggregates acoustic frames into discrete speech representations that closely mirror ground-truth token lengths across both LLM backbones. Specifically, the average length ratio ($M_{\text{CIF}} / M$, where $M_{\text{CIF}}$ and $M$ denote the number of emitted ACIF tokens and target text tokens, respectively) remains tightly bounded between $0.994$ and $0.996$ with minimal variance ($\sigma \le 0.033$). This demonstrates that the learned $\alpha$-accumulation accurately triggers at lexical boundaries without systematic under- or over-generation. Remarkably, $82\%\text{--}84\%$ of test sequences achieve an exact length match with the reference token sequence, and over $97\%$ fall within a tight $\pm 1$ token tolerance ($>99\%$ within $\pm 2$ tokens). Furthermore, this length calibration is strictly preserved throughout Stage~2 fine-tuning under both KD and CE objectives.

Crucially for computational efficiency, ACIF achieves an average dynamic downsampling rate of $4.45\times$ over the projected speech representations. Compounded with the initial $4\times$ reduction from the projector's convolutional layers, this yields an effective overall compression ratio of $\sim 17.8\times$ ($4 \times 4.45$) relative to the speech encoder's output. In contrast, SpeechMapper relies on a fixed $8\times$ reduction via consecutive frame averaging and strided convolutions. ACIF thus achieves more than double the temporal compression of prior work, substantially shortening the sequence length injected into the frozen LLM and mitigating quadratic self-attention overhead during inference.

\begin{table}[t]
    \scriptsize
    \centering
    \newcommand{\std}[1]{{\color{gray}\scriptsize$\pm$#1}}
    \setlength{\tabcolsep}{2.2pt}
    \setlength{\aboverulesep}{0.9pt}
    \setlength{\belowrulesep}{0pt}
    \resizebox{\columnwidth}{!}{
    \begin{tabular}{@{}ll ccccc@{}}
        \toprule
        & \textbf{Stage} & \textbf{Comp.} & \textbf{Ratio} & \textbf{Exact (\%)} & \textbf{$\le \pm$1 (\%)} & \textbf{$\le \pm$2 (\%)} \\
        \midrule
        \multirow{3}{*}{\rotatebox[origin=c]{90}{\scriptsize Llama 3.1}} 
        & Stage 1      & 4.45$\times$ & \textbf{0.996}\std{0.032} & \textbf{84.1} & \textbf{97.8} & 99.3 \\
        & Stage 2 [CE] & 4.45$\times$ & 0.995\std{0.033}          & 82.1          & 97.1          & 99.2 \\
        & Stage 2 [KD] & 4.45$\times$ & 0.994\std{0.032}          & 82.7          & 97.1          & 99.2 \\[2ex]
        \midrule
        \multirow{3}{*}{\rotatebox[origin=c]{90}{\scriptsize Qwen 2.5}} 
        & \rule{0pt}{2ex}Stage 1      & 4.45$\times$ & 0.996\std{0.031}          & 83.8          & 97.7          & \textbf{99.4} \\
        & Stage 2 [CE] & 4.45$\times$ & 0.995\std{0.033}          & 82.8          & 97.4          & 99.3 \\
        & Stage 2 [KD] & 4.45$\times$ & 0.995\std{0.033}          & 83.2          & 97.3          & 99.2 \\[2ex]
        \bottomrule
    \end{tabular}
    }
    \caption{Acoustic sequence compression and length alignment accuracy on LibriSpeech \texttt{test-clean}. \textbf{Comp.} denotes the dynamic CIF compression factor ($N' / M_{\text{CIF}}$). \textbf{Ratio} indicates the mean length ratio ($M_{\text{CIF}} / M \pm \sigma$). \textbf{Exact}, \textbf{$\le \pm$1}, and \textbf{$\le \pm$2} report the percentage of utterances matching the target transcript length within 0, 1, and 2 token tolerances. Best overall results are in bold.}
    \label{tab:length_matching_results}
\end{table}

\section{Conclusion}

In this paper, we introduced ACIF, a highly efficient framework that equips text-centric LLMs with zero-shot speech capabilities. By dynamically compressing acoustic frames into exact discrete token lengths via DTW-guided alignments, ACIF enhances both computational efficiency and downstream performance while eliminating the need for full LLM supervision. Our Stage 1 training establishes a robust acoustic-to-semantic bridge using lightweight distance metrics, entirely bypassing the LLM forward pass. For optional Stage 2 fine-tuning, we proposed a single-layer knowledge distillation objective that performs competitively with full-model cross-entropy supervision at a fraction of the memory cost. Ultimately, ACIF consistently outperforms parameter-efficient baselines on ASR and ST benchmarks, providing a lightweight, robust alternative for the future development of SpeechLLMs.

\section*{Limitations}
While ACIF demonstrates strong zero-shot capabilities, several limitations remain. First, our current training and evaluation pipelines are primarily English-centric. Future work should investigate the framework's scalability to a broader range of low-resource and typologically diverse languages. Second, our empirical evaluation focuses exclusively on Automatic Speech Recognition and Speech Translation. Exploring the applicability of ACIF to broader downstream tasks, such as spoken question answering or acoustic reasoning, remains an open area for investigation. Third, due to computational constraints, all models were trained using a single random seed. Although training converged reliably across configurations, reporting variance and confidence intervals across multiple seeds would provide a more complete assessment of performance stability. Finally, mapping continuous speech frames directly to discrete text token embeddings runs the risk of discarding valuable prosodic and paralinguistic cues inherent to spoken language. Future research should systematically analyze this information loss and explore alignment mechanisms that retain expressive acoustic features without compromising semantic alignment.

\section*{Acknowledgments}
The research presented in this paper was conducted as part of VOXReality project\footnote{\texttt{\url{https://voxreality.eu/}}}, which was funded by the European Union Horizon Europe program under grant agreement No 101070521.

This work used the Dutch national e-infrastructure with the support of the SURF Cooperative using grant no. EINF-16552.

\bibliography{custom}

\appendix

\section{Evaluation Prompts}
\label{sec:prompt}
For our evaluations, we format the inputs according to the specific chat templates of the underlying LLMs (e.g., Llama-3.1 or Qwen-2.5). The generalized prompt structure consists of a system message establishing the assistant's role, followed by a user message containing the projected acoustic embeddings and a task-specific instruction. 

For ASR, we follow the instruction style of SpeechMapper to prompt verbatim transcription. For ST, the user instruction consists of a target-language-specific instruction which is a translation of "Can you translate the Speech content into [German/Spanish/French/Italian/Chinese] text?", followed by a formatting suffix: \textit{``Output only the translation and nothing else.''}. Table~\ref{tab:evaluation_prompts} details the exact system and user instructions used across all evaluations.

\begin{table*}[ht]
\centering
\resizebox{\textwidth}{!}{
\begin{tabular}{@{}llp{10cm}@{}}
\toprule
\textbf{Task} & \textbf{System Prompt} & \textbf{User Instruction (Appended after Speech Embeddings)} \\ 
\midrule
\textbf{ASR} & You are a helpful ASR transcription assistant. & Repeat the previous text between the quotes in its entirety just once and nothing else. Do not repeat the text multiple times or correct the text or add punctuation. End the text if you notice a phrase or a text is getting repeated. Ignore the words that do not make any sense. \\ 
\addlinespace
\midrule
\addlinespace
\textbf{ST (de)} & You are a helpful translation assistant. & Können Sie den Inhalt der Rede in den deutschen Text übersetzen?\newline Output only the translation and nothing else. \\ 
\addlinespace
\textbf{ST (es)} & You are a helpful translation assistant. & ¿Puedes traducir el contenido del discurso al texto en español?\newline Output only the translation and nothing else. \\ 
\addlinespace
\textbf{ST (fr)} & You are a helpful translation assistant. & Pouvez-vous traduire le contenu du discours en texte français ?\newline Output only the translation and nothing else. \\ 
\addlinespace
\textbf{ST (it)} & You are a helpful translation assistant. & Puoi tradurre il contenuto del discorso in testo italiano?\newline Output only the translation and nothing else. \\ 
\addlinespace
\textbf{ST (zh)} & You are a helpful translation assistant. & \begin{CJK*}{UTF8}{gbsn}你能把演讲内容翻译成中文吗？\end{CJK*}\newline Output only the translation and nothing else. \\ 
\bottomrule
\end{tabular}
}
\caption{System prompts and user instructions used for zero-shot ASR and ST evaluation. The acoustic embeddings are inserted directly before the user instruction to form the final user message.}
\label{tab:evaluation_prompts}
\end{table*}

\section{Cosine is as Effective as KL}
\label{sec:cosine}
Computing the standard KL divergence requires projecting hidden states through the LLM's unembedding matrix and applying a softmax operation, which introduces notable computational overhead during training. As a more efficient alternative, we experiment with defining our distillation objective, $\mathcal{L}_{\text{KD}}$, as the simple cosine distance between the intermediate hidden states of the speech representations $Z$ and the text embeddings $E$. Because the cosine distance is strictly bounded between 0 and 2, yielding intrinsically smaller numerical values than standard CE or KL losses, we scale this objective by a factor of $\lambda_{\text{kd}} = 10.0$. As demonstrated in Table \ref{tab:kd_objective}, the cosine distance serves as a highly effective alternative to KL divergence. It achieves identical average COMET scores (77.8) across the Europarl-ST translation directions, while incurring only a marginal performance degradation of 0.1 WER on the LibriSpeech test-clean split.
\begin{table}[H]
    \centering
    \begin{tabular}{@{}l cc@{}}
        \toprule
        \textbf{Objective} & \textbf{LS clean} & \textbf{Europarl-ST} \\
        \midrule
        Cosine & 4.2 / 2.3 & \textbf{77.8} \\
        KL & \textbf{4.1 / 2.2} & \textbf{77.8} \\
        \bottomrule
    \end{tabular}
    \caption{Performance comparison between Cosine and KL distillation objectives on LibriSpeech (WER/CER) and Europarl-ST (COMET).}
    \label{tab:kd_objective}
\end{table} 

\end{document}